\documentclass[10pt,twocolumn,letterpaper]{article}

\usepackage{3dv}
\usepackage{times}
\usepackage{epsfig}
\usepackage{graphicx}
\usepackage{amsmath}
\usepackage{amssymb}
\usepackage{gensymb}
\usepackage{subcaption}
\usepackage{mwe}

\usepackage[pagebackref=true,breaklinks=true,letterpaper=true,colorlinks,bookmarks=false]{hyperref}

\newcommand\blfootnote[1]{%
  \begingroup
  \renewcommand\thefootnote{}\footnote{#1}%
  \addtocounter{footnote}{-1}%
  \endgroup
}

\threedvfinalcopy 

\def\threedvPaperID{212} 

\ifthreedvfinal\fi
\begin{document}

\newcommand{\vova}[1]{{\bf\textcolor{red}{Vova: #1}}}

\title{Self-Supervised Learning of Point Clouds via Orientation Estimation}

\author{Omid Poursaeed$^{*1}$ \qquad
 Tianxing Jiang$^{*1}$ \qquad
 Han Qiao$^{*1}$ \qquad
 Nayun Xu$^{1}$ \qquad
 Vladimir G. Kim$^{2}$ \\ \\
  	$^1${Cornell University}\qquad
    $^2${Adobe Research}}
\maketitle
\begin{abstract}
Point clouds provide a compact and efficient representation of 3D shapes. While deep neural networks have achieved impressive results on point cloud learning tasks, they require massive amounts of manually labeled data, which can be costly and time-consuming to collect. In this paper, we leverage 3D self-supervision for learning downstream tasks on point clouds with fewer labels. A point cloud can be rotated in infinitely many ways, which provides a rich label-free source for self-supervision. We consider the auxiliary task of predicting rotations that in turn leads to useful features for other tasks such as shape classification and 3D keypoint prediction. Using experiments on ShapeNet and ModelNet, we demonstrate that our approach outperforms the state-of-the-art. Moreover, features learned by our model are complementary to other self-supervised methods and combining them leads to further performance improvement.    
   
\end{abstract}

\section{Introduction}
\blfootnote{* Indicates equal contribution}
As a concise and expressive form of representing 3D shapes, point clouds are ubiquitous in various 3D vision tasks. 
Massive point cloud datasets have become available with the development of 3D scanning devices and techniques. 
ShapeNet~\cite{chang2015shapenet}, ModelNet~\cite{wu20153d} and ScanNet~\cite{dai2017scannet} are instances of such datasets. Models such as PointNet~\cite{qi2017pointnet} have achieved impressive results on these datasets. However, manually annotating these large amounts of data is costly and impractical to scale. 
This raises the need for methods that can learn from unlabelled data to reduce the number of annotated samples required for learning. Recently, self-supervised learning have shown promising performance on several computer vision tasks. Self-supervision is a learning framework in which a supervised signal for a pretext task is created automatically, in an effort to learn representations that are useful for solving real-world downstream tasks. 

A simple form of self-supervision is predicting rotations \cite{gidaris2018unsupervised}. This approach has achieved state-of-the-art results on image classification~\cite{kolesnikov2019revisiting}. 
In this paper we propose the first self-supervised approach based on rotations of 3D data. 
Among various 3D representations, we focus on point clouds as an efficient representation in which rotations can be easily applied via matrix multiplication. 
%
Manually annotating point clouds to provide information such as category labels or keypoints is cumbersome and costly. 
%
%
Notably, most public 3D models have a \textit{canonical} orientation, or at least a known \textit{up} axis. We can leverage this orientation information to provide a label-free supervision signal for downstream tasks. More specifically, we can arbitrarily rotate an input point cloud and train a model to recover the rotation angle. 
The auxiliary task of predicting these rotations requires high-level understanding of 3D shapes, since canonical orientation of the object implicitly encodes important information about its functionality~\cite{Fu2008upright}.

We demonstrate that a model trained for this proxy task generates semantic features useful for target tasks such as classification and keypoint prediction. 
We show that simple models such as SVMs can be trained on these features and achieve strong performance even when trained on a small subset of the training data. Hence, our approach facilitates learning with fewer labels, and obviates the need for manually annotating massive datasets. Finally, we discuss how our approach can be combined with other self-supervised models to achieve further performance gain, indicating that our
learned features provide complementary information to other methods.  

\section{Related Work}

\subsection{Deep Learning on Geometry}

There has been a recent surge of interest on learning-based approaches on geometric data~\cite{brock2016generative,choy20163d,rezende2016unsupervised,wu2016learning,wu20153d}. These models are able to encode rich prior information about the space of 3D shapes which helps to resolve ambiguities in the input. 
Typical convolutional architectures require highly regular input data formats such as image grids or 3D voxels. 
Unlike images, geometry usually does not have an underlying grid structure, requiring new building blocks replacing convolution and pooling or adaptation to a grid. 
As a simple way to overcome this issue, view-based~\cite{su2015multi,wei2016dense,poursaeed2018deep}
and volumetric representations~\cite{klokov2017escape,maturana2015voxnet,tatarchenko2017octree,wu20153d} or their combination~\cite{qi2016volumetric} place geometric data onto a grid. 
Voxel representations are a straightforward generalization of pixels to the 3D case. However, the memory footprint of these representations grows cubically with resolution. Data adaptive representations such as octrees~\cite{riegler2017octnet,tatarchenko2017octree} mitigate this issue, yet they lead to complex implementations and are still limited to relatively small voxel grids. 

Point clouds provide a more efficient representation by only focusing on surface points, yielding a tensor with a fixed number of 3D coordinates~\cite{achlioptas2018learning,fan2017point}. 
Point clouds as an input modality present a unique set of challenges when building a network architecture. 
The grid structure is not available in raw point clouds as unordered sets of points. 
To address this issue, Qi et al. propose PointNet~\cite{qi2017pointnet} which respects permutation invariance of point clouds by using a max-pooling operation to form a single feature vector representing the global context. 
PointNet++~\cite{qi2017pointnet++} extends PointNet by recursively applying it on a nested partitioning of the input point set.
Kernel correlation and graph pooling are proposed for improving PointNet-like methods in~\cite{shen2018mining}. 
PointCNN~\cite{li2018pointcnn} leverages spatially-local correlations using a $\chi$-Conv operator that weights and permutes input points and features before they are processed by a typical convolution. 
PointConv~\cite{wu2019pointconv} and Relation-Shape CNN~\cite{liu2019relation} also focus on local structures of point clouds and further improve the quality of captured features.
Recently, graph convolutional networks have shown promising results on learning tasks on point clouds.  
Dynamic Graph CNN~\cite{wang2019dynamic} applies a graph convolution to edges of the k-nearest neighbor graph of the point clouds, which is dynamically recomputed in the feature space after each layer. While the aforementioned models achieve great performance on various tasks, they require large number of annotated samples for training. Hence, there is a need for 
efficient methods with less supervision.     

Learning-based approaches have been applied to several other geometric representations such as meshes~\cite{kanazawa2018learning,ranjan2018generating,wang2018pixel2mesh,poursaeed2020neural}, implicit functions~\cite{mescheder2019occupancy,park2019deepsdf,xu2019disn,poursaeed2020coupling} and structured models~\cite{li2017grass,zou20173d,zhu2018scores}. We refer the interested reader to~\cite{cao2020comprehensive} for a survey on geometric deep learning.   

\subsection{Unsupervised and Self-Supervised Learning}

The goal of unsupervised learning is to learn representations from raw data without the use of labels such that they are easily adaptable for other tasks. 
A common approach in unsupervised learning is leveraging generative models such as Variational AutoEncoders (VAEs)~\cite{kingma2014semi,kingma2013auto,rezende2014stochastic}, Generative Adversarial Networks (GANs)~\cite{goodfellow2014generative,huang2017stacked,radford2015unsupervised,kiapour2020generating} and autoregressive models~\cite{gregor2014deep,van2016conditional,van2016pixel}.   
Another prominent paradigm is self-supervision, which defines an annotation-free pretext task in order to provide a surrogate supervision signal for feature learning. Example pretext tasks include colorizing gray scale images~\cite{larsson2016learning,zhang2016colorful,zhang2017split}, image inpainting~\cite{pathak2016context}, noise-as-targets~\cite{bojanowski2017unsupervised}, 
generation~\cite{jenni2018self,ren2018cross}, geometry~\cite{dosovitskiy2015discriminative}, counting~\cite{noroozi2017representation} and clustering of images~\cite{caron2018deep,liao2016learning,yang2016joint}. 
Other methods use additional information such as temporal data to learn features. 
Typical proxy tasks are based on temporal-context ~\cite{misra2016shuffle,wei2018learning,sermanet2018time}, spatio-temporal cues~\cite{isola2015learning,gao2016object,wang2017transitive}, foreground-background segmentation via video segmentation~\cite{pathak2017learning}, 
optical-flow~\cite{gan2018geometry,mahendran2018cross},  future-frame synthesis~\cite{srivastava2015unsupervised}, audio prediction from video~\cite{de1994learning,owens2016visually}, audio-video alignment~\cite{arandjelovic2017look}, ego-motion estimation~\cite{jayaraman2016slow}, slow feature analysis with higher order temporal coherence~\cite{jayaraman2016slow}, transformation between frames~\cite{agrawal2015learning} and patch tracking in videos~\cite{wang2015unsupervised}.

Several papers consider predicting the spatial relation between two randomly chosen image
patches~\cite{doersch2015unsupervised,nathan2018improvements,noroozi2016unsupervised,noroozi2018boosting}. An extension of this idea to 3D representations is presented in~\cite{sauder2019context}. 
%
%
Our work bears a resemblance to~\cite{gidaris2018unsupervised} which randomly rotates an image by one of four possible angles and let the model predict the rotation. 
 ~\cite{oord2018representation} proposes to predict future patches in representation space via autoregressive predictive coding.
An overview and comparison of various 2D self-supervised learning algorithms is provided in~\cite{kolesnikov2019revisiting,asano2019acritical}. 

Recently, various methods have been proposed for unsupervised learning of point clouds. Autoencoder-based approaches learn features through reconstructing the point cloud data~\cite{achlioptas2018learning,gadelha2018multiresolution,yang2018foldingnet,zhao20193d}. Generative models learn representations in the process of generating plausible point clouds~\cite{li2018point,sun2020pointgrow,thabet2020self,wu2016learning}. Concurrent to our work, various pretext tasks have been proposed for self-supervision on point clouds. Sauder et al. propose to learn features by reconstructing point clouds whose parts have been randomly rearranged~\cite{sauder2019context}. 
Zhang et al. train deep graph neural networks to solve two proxy tasks, part contrasting and object clustering~\cite{zhang2019unsupervised}.  
A multi-task learning framework is proposed in~\cite{hassani2019unsupervised} which learns features by optimizing three different tasks including clustering, prediction, and reconstruction. 
Our approach requires more holistic understanding of shapes, and is relatively easy to implement. As we demonstrate in experiments, our model outperforms several competing methods, and can also be combined with them to achieve further performance gain. 
%
\begin{figure*}[t]
  \includegraphics[width=\linewidth]{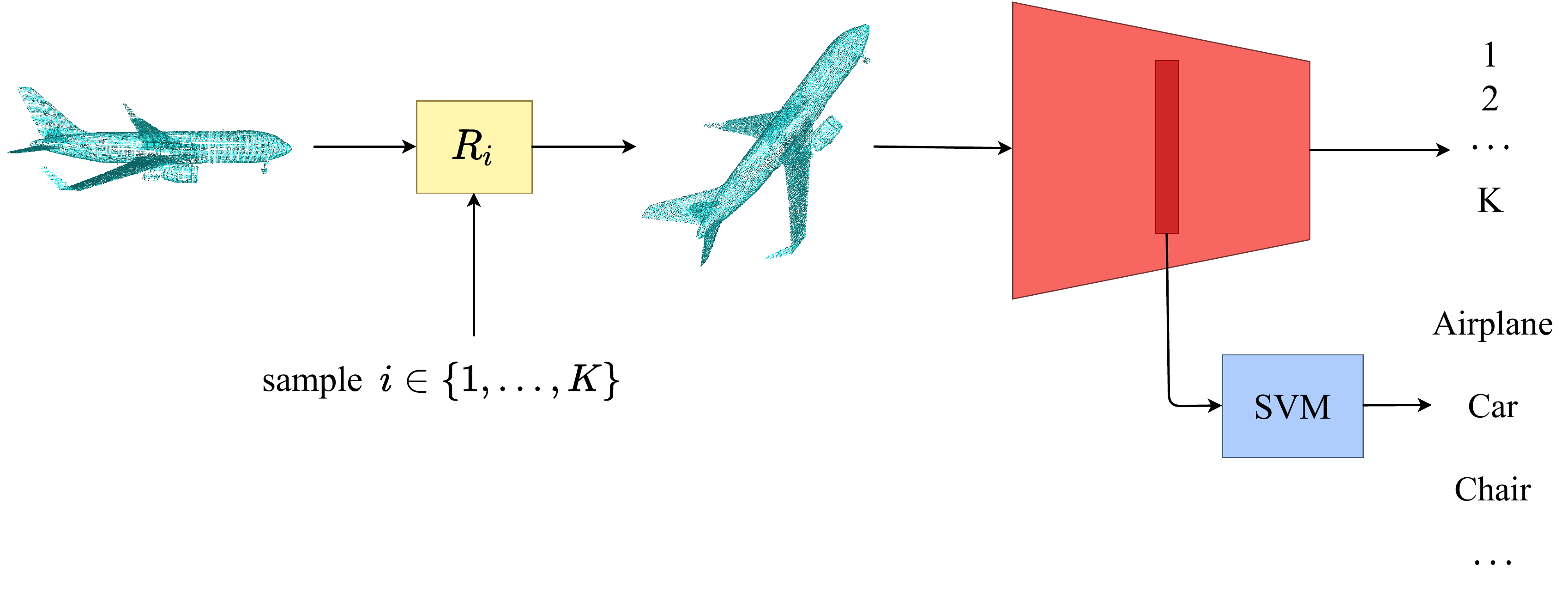}
\caption{Model architecture. A rotation angle is randomly sampled from the set of possible angles and applied to the input point cloud. The rotation prediction model is trained to infer the rotation angle from the rotated point cloud. Features from the rotation prediction model are passed to a linear SVM which is trained to infer the shape category. 
}
\label{fig:architecture}
\end{figure*}

\section{Method}
\label{sec:method}
Inspired by the success of self-supervised learning on images, we propose a self-supervised method to learn representations of point clouds. We consider the proxy task of rotation prediction. 
A set of $K$ rotations $R = \{R_1, R_2, \dots , R_K\}$ is used, where $R_i(X)$ rotates upward direction of the input point cloud $X$ to the $i^\text{th}$ angle. We train a classifier $F(\cdot)$ to predict the rotation applied to the input point cloud. 
The model takes a transformed point cloud $R_i(X)$, with $R_i$ randomly selected from $R$, and outputs a probability distribution over all possible rotations. We train the model using cross-entropy loss in a supervised manner.

Figure~\ref{fig:architecture} illustrates our architecture. Our models are based on PointNet~\cite{qi2017pointnet} and DGCNN~\cite{wang2019dynamic} in which the final layers are modified for rotation classification or regression. 
The classifier outputs $K$ probabilities and the regression model outputs four real numbers (three for the rotation axis and one for the rotation angle). 
The features from the rotation prediction model are then used for several target tasks including classification and keypoint prediction. 
ShapeNet~\cite{chang2015shapenet}, a large scale dataset of 3D shapes, is used to train the rotation estimation model. 3D models in ShapeNet are annotated with upright and front orientations, allowing us to align them to a canonical orientation. 
\begin{figure*}
\begin{center}
\begin{subfigure}{.45\textwidth}
  \centering
  \includegraphics[width=0.75\linewidth]{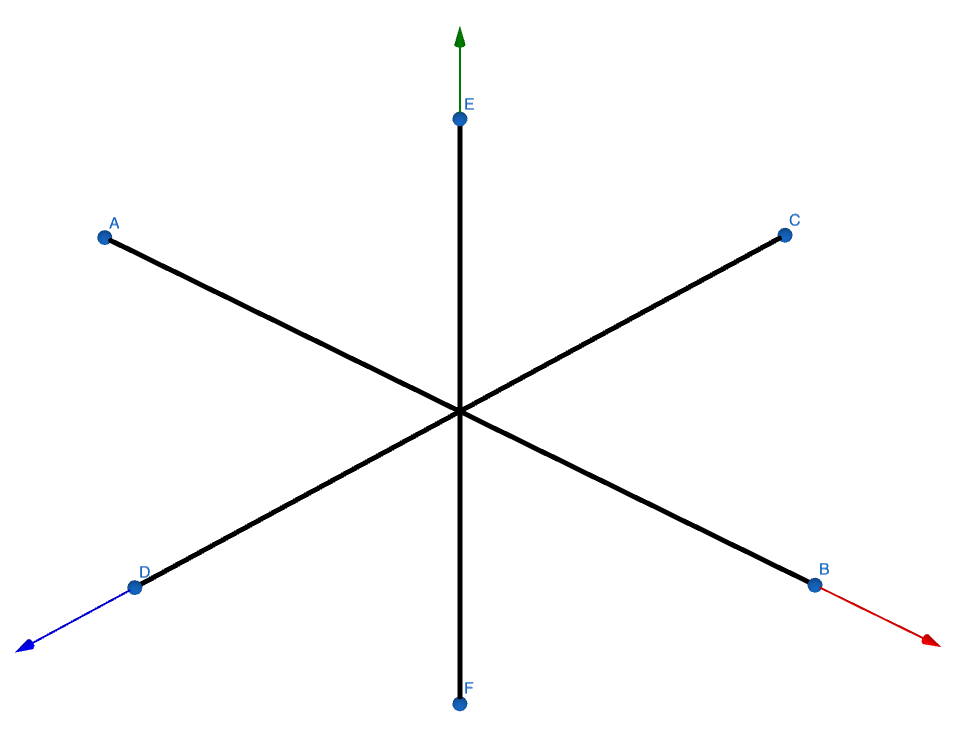}
  \caption{$K=6$: directions are on the $\pm x,\pm y,\pm z$ axes.}
  \label{fig:sfig1}
\end{subfigure}
\hspace{5pt}
\begin{subfigure}{.4\textwidth}
  \centering
  \includegraphics[width=0.75\linewidth]{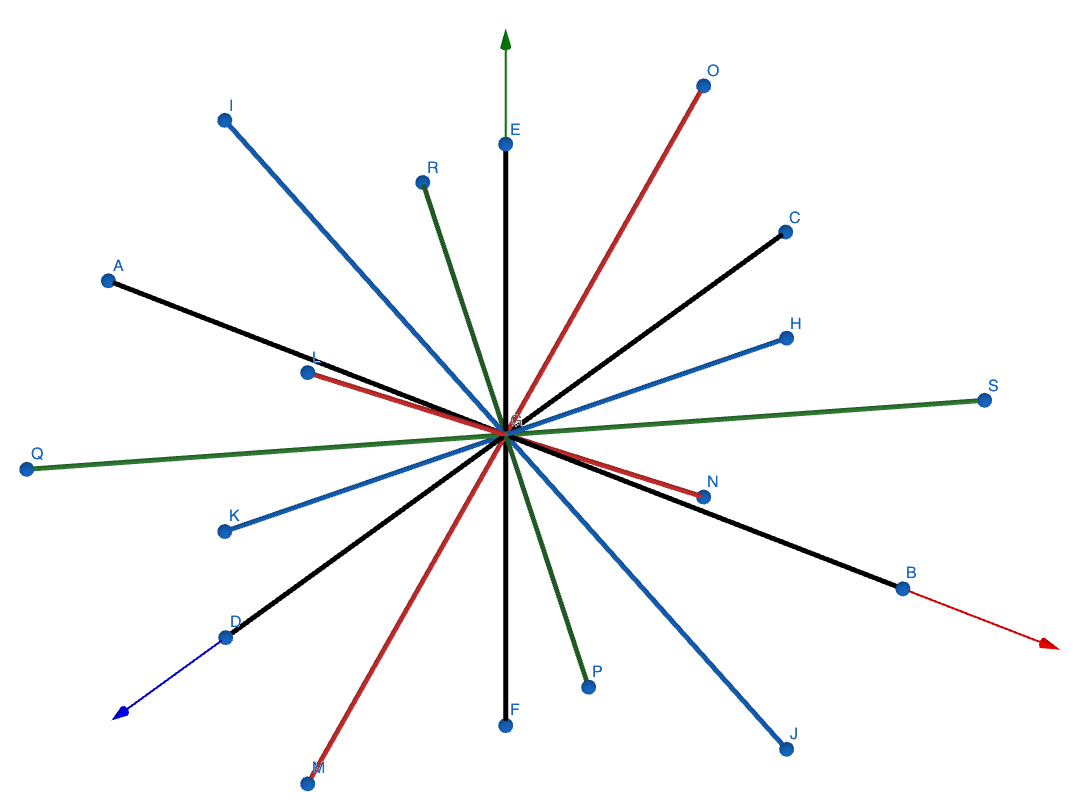}
  \caption{$K=18$: directions on the xy-plane are blue; directions on the xz-plane are green; directions on the yz-plane are red.}
  \label{fig:sfig2}
\end{subfigure}
\hspace{5pt}
\begin{subfigure}{.5\textwidth}
  \centering
  \includegraphics[width=0.76\linewidth]{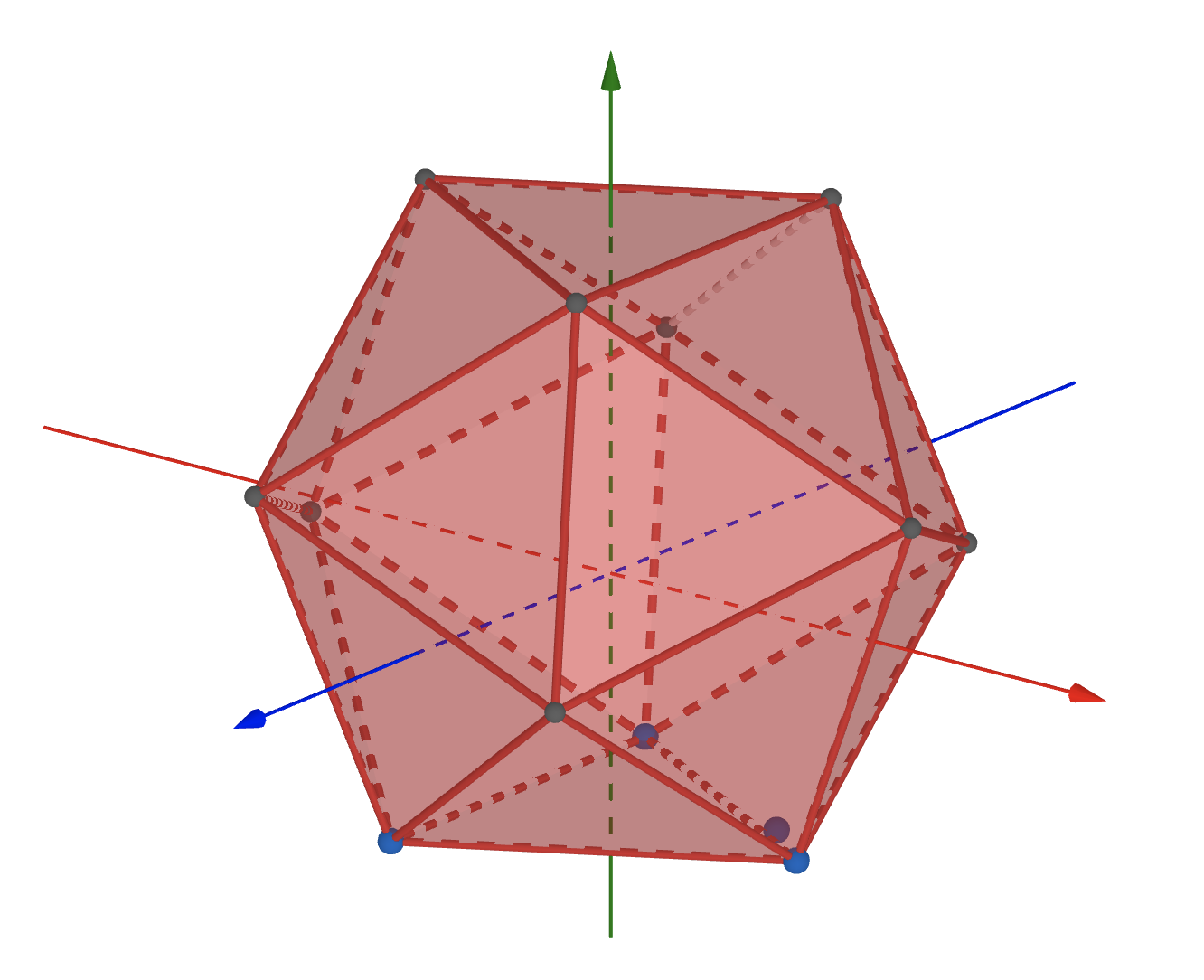}
  \caption{$K=32$: directions are towards the vertices and face centers of the regular icosahedron.}
  \label{fig:sfig3}
\end{subfigure}
\hspace{5pt}
\begin{subfigure}{.4\textwidth}
  \centering
  \includegraphics[width=0.75\linewidth]{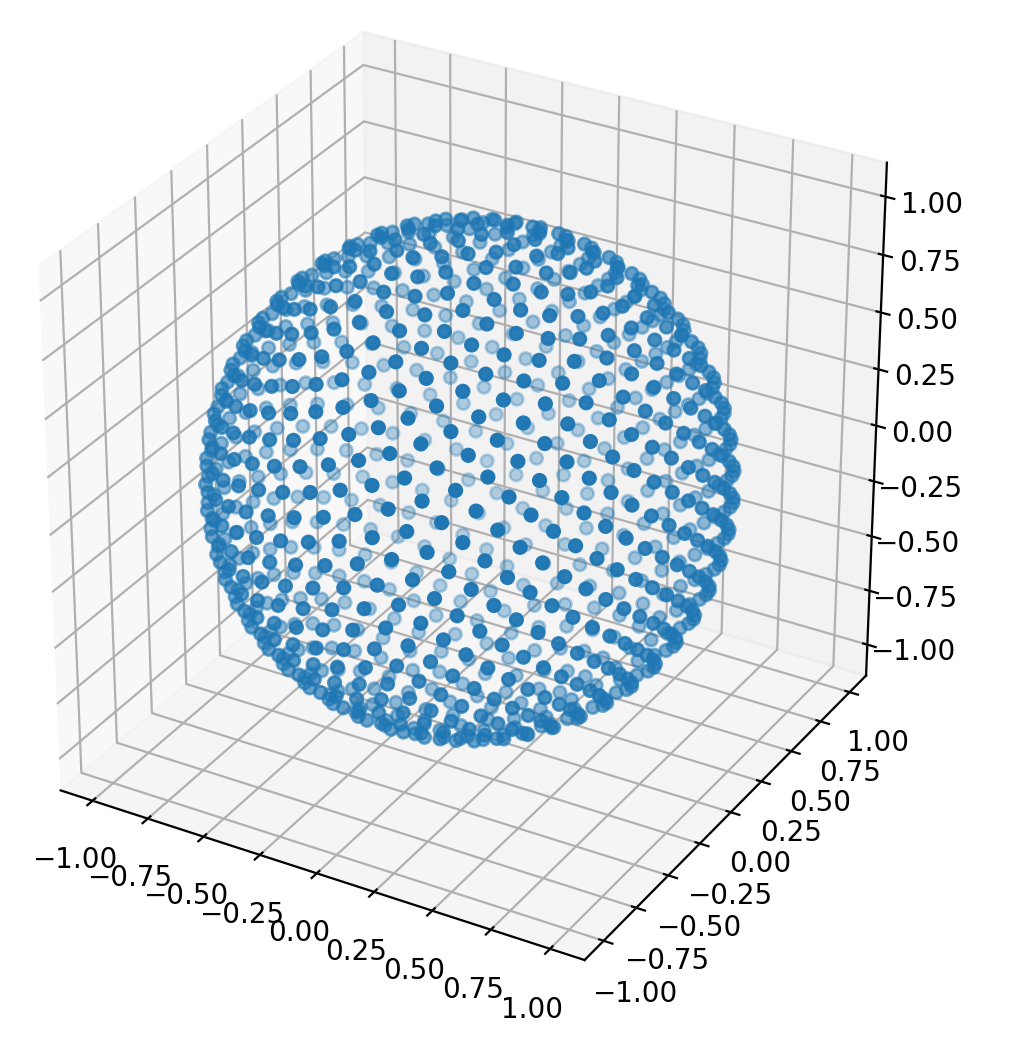}
  \caption{$K=54, 100$: the points conform to the golden spiral sunflower distribution.}
  \label{fig:sfig4}
\end{subfigure}
\end{center}
   \caption{Illustration of the rotation angles. The object is placed at the origin, and its upward direction is rotated to one of the $K$ marked points.
   }
\label{fig:angles}
\end{figure*}

Number of angles $K$ is a hyper-parameter. 
Using a small $K$ gives a high accuracy on rotation prediction while might not yield the best features for downstream tasks since predicting a small number of rotations is a simple task. On the other hand, a very large $K$ gives a low accuracy on rotation prediction as angles become closer to one other. Table \ref{tab:rot accuracy shapenet} shows the rotation classification accuracy for different values of $K$. As we observe, the accuracy decreases for larger $K$ values.

\begin{table}
\begin{center}
\begin{tabular}{|c|c|}
\hline
Number of rotation angles $K$ & Prediction accuracy \\
\hline\hline
6 & 99.8\% \\
18 & 89.0\% \\
32 & 90.3\% \\
54 & 67.8\% \\
100 & 1.6\% \\
\hline
\end{tabular}
\end{center}
\caption{Rotation classification accuracy on ShapeNet based on the number of angles $K$.}\label{tab:rot accuracy shapenet}
\end{table}

For a given choice of $K$ we need to select the angles as uniformly as possible to avoid bias in learning. Ideally, we can consider a regular polyhedron with $K$ vertices. However, regular polyhedra do not exist for all values of $K$, hence we resort to approximate solutions for custom $K$ values. 
We consider a range of small to large values for $K$ including $6, 18, 32, 54$ and $100$.  
For six rotation angles, we choose the six directions of the $\pm x, \pm y, \pm z$ axes. 
For $K=18$, we take directions of the six axes and the twelve angle bisectors for each two consecutive axes. 
For $K=32$, a regular icosahedron is centered at the origin, and directions towards its 12 vertices and 20 face centers are chosen.
For $K=54, 100$, we use the golden spiral sunflower distribution, where the golden ratio is $\frac{1+\sqrt{5}}{2}$. Figure ~\ref{fig:angles} illustrates the distribution of angles for various $K$ values. 

Alternatively, we can train a model to regress the angles instead of classifying them. We consider two rotation representations for the regression task: axis-angle and a continuous 6D representation presented in~\cite{zhou2019continuity}.  
For the axis-angle representation, we first randomly select an axis by sampling coordinates $x, y, z \sim \mathcal{N}(0, 1)$ and normalizing the resulting vector $(x, y, z)^T$ to have a unit norm. It can be shown that the resulting vectors are uniformly distributed on the unit sphere. We then sample a random angle, and rotate the input point cloud using the sampled axis and angle. A regression model is the trained to regress the axis of rotation and the rotation angle with respect to this axis. 
We train the model with $L_2$ loss using a linear combination of axis and angle prediction losses with equal weights. 

We also consider the representation proposed by Zhou et al.~\cite{zhou2019continuity}. They demonstrate that for 3D rotations all representations are discontinuous in the real Euclidean spaces of four or fewer dimensions. 
They present continuous representations in 5D and 6D which are more suitable for learning. We use the 6D representation and consider a similar setup as their experiment on pose estimation for 3D point clouds. The network consists of a simplified PointNet structure for feature extraction and an MLP to produce the 6 dimensional rotation representation. A mapping function transforms this representation to \textit{SO}(3). Note that while they estimate the relative rotation between two point clouds, we predict the rotation of a single point cloud with respect to a canonical orientation.  

Finally, we use the trained rotation estimation model to extract features for downstream tasks. These features are passed to a linear SVM for shape classification. In the keypoint prediction task, we modify final layers of the angle prediction model and fine-tune it for keypoint regression. We train the rotation estimation and target models on ShapeNet and ModelNet respectively.

Note that our method assumes that the shapes have been consistently aligned, but this is a common assumption in online stock repositories. 
We can extend our approach to predict the relative rotation between point clouds either in a supervised or unsupervised manner. 
In the supervised case we use ground truth orientations, and it is quite similar to our current setup. In the unsupervised case we can align a point cloud to a rotated version of itself. We can estimate the relative rotation, apply it to one of the point clouds and use a chamfer loss to penalize misalignment. The latter case enables learning even in the absence of canonical orientations. However, the learned features might not be as transferable as current features. We will further explore this direction in future work.  

\begin{figure*}
\begin{center}
\begin{subfigure}{.49\textwidth}
  \centering
  \includegraphics[width=\linewidth]{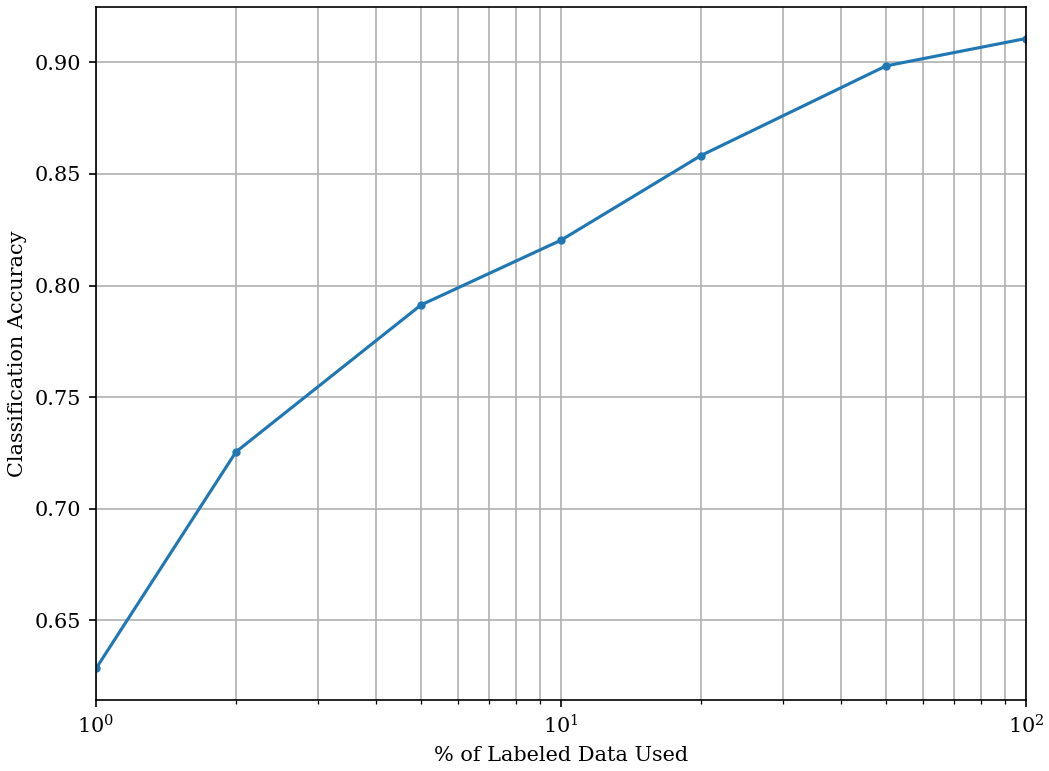}
  \caption{$K=18$}
  \label{fig:ssvm1}
\end{subfigure}
\begin{subfigure}{.49\textwidth}
  \centering
  \includegraphics[width=\linewidth]{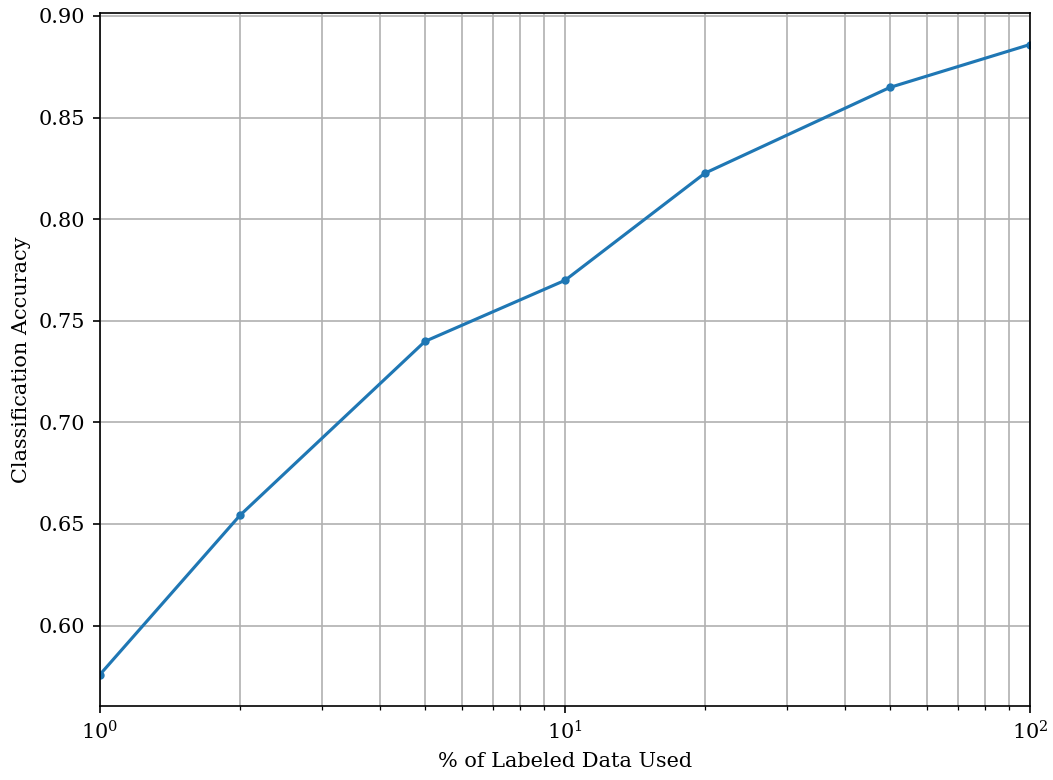}
  \caption{$K=32$}
  \label{fig:ssvm2}
\end{subfigure}
\end{center}
\caption{Linear SVM's classification accuracy on ModelNet-40 for different sizes of training set. }
\label{fig:portion}
\end{figure*}

\section{Experiments}  

We demonstrate that features learned by the rotation prediction model are useful for downstream tasks. We consider two tasks: shape classification and 3D keypoint prediction. 

\subsection{Shape Classification}
We consider the task of shape classification on ModelNet~\cite{wu20153d}. 
Using the rotation prediction model trained on ShapeNet, we generate features for ModelNet-40's train split. We use the global activations after the max pooling layers of PointNet and DGCNN as we found them to yield the best performance. 
A linear SVM classifier is trained on these activations for the task of predicting the shape category. This experiment evaluates the learned features in a transfer learning task, demonstrating their generalizability. Figure ~\ref{fig:architecture} illustrates the architecture. Note that the rotation estimation model is trained on the ShapeNet dataset while the linear SVM is trained on features from ModelNet.

\begin{table}
\begin{center}
\begin{tabular}{c c c}
\hline\hline
Previous work \\
\hline\hline
VConv-DAE~\cite{sharma2016vconv} & 75.50\% \\ 
3D-GAN~\cite{wu2016learning} & 83.30\% \\ 
Latent-GAN~\cite{achlioptas2018learning} & 85.70\% \\ 
FoldingNet~\cite{yang2018foldingnet} & 88.40\% \\ 
VIP-GAN~\cite{han2019view} & 90.19\% \\ 
Context Prediction (DGCNN)~\cite{sauder2019context} & 90.64\% \\ 
Context Prediction (PointNet)~\cite{sauder2019context} & 87.31\% \\ 
\hline
\hline
Ours (DGCNN) \\
\hline
\hline
6 angles & 90.06\% \\ 
18 angles & {\bf 90.75}\% \\ 
32 angles & 89.41\% \\ 
\hline
\hline
Ours (PointNet) \\
\hline
\hline
6 angles & 87.5\% \\ 
18 angles & 88.5\% \\ 
32 angles & 88.6\% \\ 
\hline
\hline
Ours + Context Prediction & {\bf 91.84\%} \\
\hline
\hline
\end{tabular}
\end{center}
\caption{Test accuracy on ModelNet-40 shape classification using the pretext task of rotation classification on ShapeNet.  
}\label{tab:svm}
\end{table}

\begin{table}
\begin{center}
\begin{tabular}{c c c}
\hline
\hline
Continuous 6D~\cite{zhou2019continuity} & 86.16\% \\
\hline
\hline
Axis-angle (PointNet) & 85.51\% \\ 
\hline
\hline
Axis-angle (DGCNN) & 83.12\% \\ 
\hline
\hline
\end{tabular}
\end{center}
\caption{Test accuracy on ModelNet-40 shape classification using the pretext task of rotation regression on ShapeNet. Different rotation representations are considered.  
}\label{tab:svm-regress}
\end{table}

Table~\ref{tab:svm} shows accuracy of the linear SVM on the test set of ModelNet-40. It also compares our model's performance with various competing methods. 
As we observe, our approach outperforms competing methods. Note that the accuracy depends on the number of angles, with $K=18, 32$ achieving the best performance. 
Using too large or too small values for $K$ leads to inferior performance as the rotation prediction task becomes too difficult or too easy. 

We can also combine our approach with other self-supervised learning methods. 
More specifically, we concatenate the features learned by our model with features extracted from the context prediction model~\cite{sauder2019context} and train the SVM classifier on the combined features. As shown in Table~\ref{tab:svm}, this leads to improved performance compared to individual models, demonstrating that our approach learns complementary features to other self-supervised methods.  Intuitively, our rotation estimation model learns global orientation information while ~\cite{sauder2019context} learns part-level information required for the task of context prediction. Both of these information sources are useful for the shape classification task, therefore combining them leads to improvements over each individual model. 

\begin{figure*}[t]
\begin{center}
\begin{subfigure}{.5\textwidth}
  \centering
  \includegraphics[width=\linewidth]{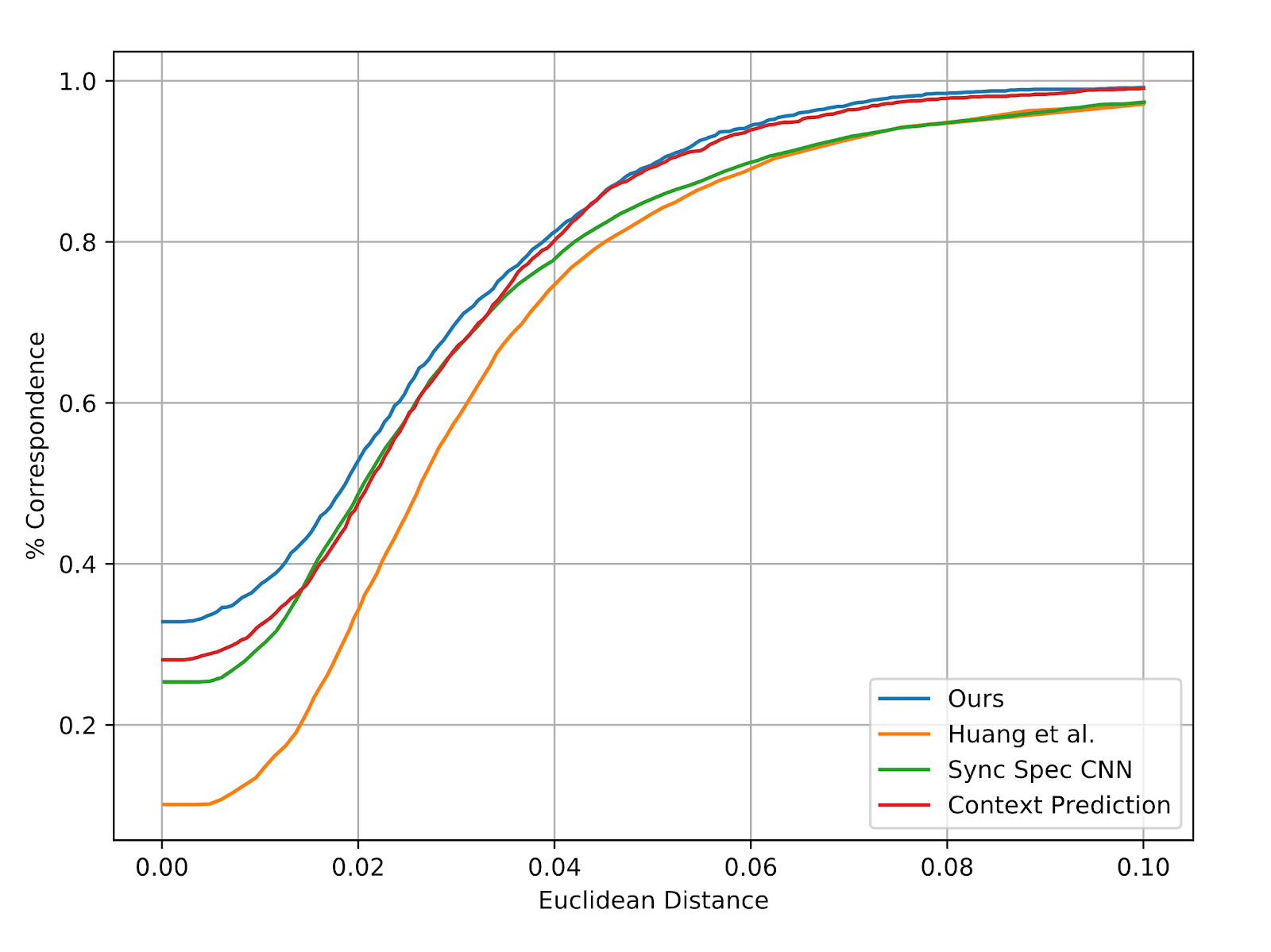}
 \end{subfigure}
 \begin{subfigure}{.49\textwidth}
   \centering
   \includegraphics[width=\linewidth]{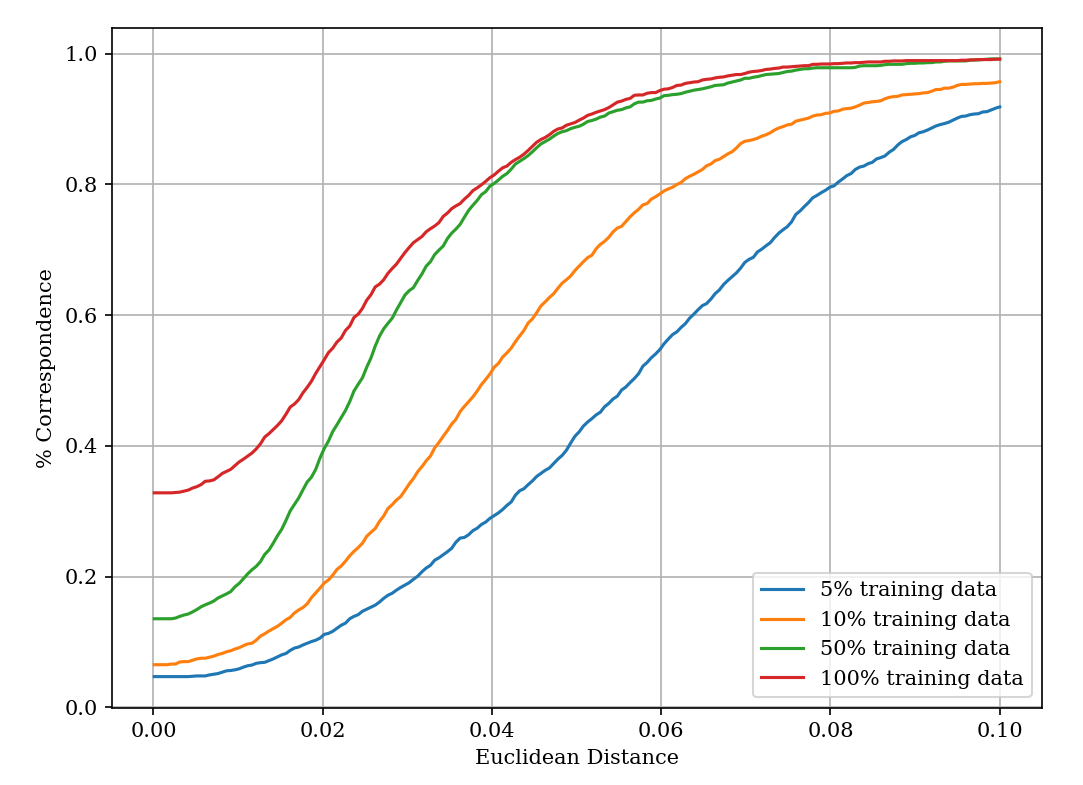}
 \end{subfigure}
\end{center}
\caption{PCK curves for the keypoint prediction task. Left: our approach outperforms competing methods for all error thresholds. Right: PCK curves for various sizes of the training set. 
}
\label{fig:pck}
\end{figure*}

We can also estimate rotations via regression instead of classification. As described in Section~\ref{sec:method}, in this approach we aim to regress the exact rotation angle rather than a discretized approximation of it via $K$ pre-selected angles. 
We consider the axis-angle representation with PointNet and DGCNN backbone architectures, as well as the 6D representation in~\cite{zhou2019continuity} with a simplified PointNet structure. 
We train the rotation regression model on ShapeNet and use it to extract features from ModelNet samples. A linear SVM classifier is trained on these features.    
Table \ref{tab:svm-regress} shows the accuracy on ModelNet-40's test set. We observe that features learned from the regression model achieve inferior performance compared with the classification model. Regressing the exact rotation angle is a challenging task. It resembles our classification setting with $K \rightarrow \infty$ which does not yield the best performance. The learned features become too task-specific and not as transferable to a shape categorization task. 

We also study dependence of our model on the size of the training set. 
We vary ModelNet-40's train set size for the linear SVM. A certain percentage of data is randomly sampled and used to train the classifier. 
We then evaluate the model's accuracy on the full test set of ModelNet-40. Figure~\ref{fig:portion} depicts the linear SVM's classification accuracy as size of the training set varies from 1\% to 100\%. 
The results indicate that we still have a relatively strong performance even when a small number of annotated samples are available. The features learned from the rotation prediction model contain information useful for the classification task. Therefore, a simple classifier trained with moderate supervision suffices to achieve high accuracy. 


\begin{figure*}
\centering
  \includegraphics[width=1.0\linewidth]{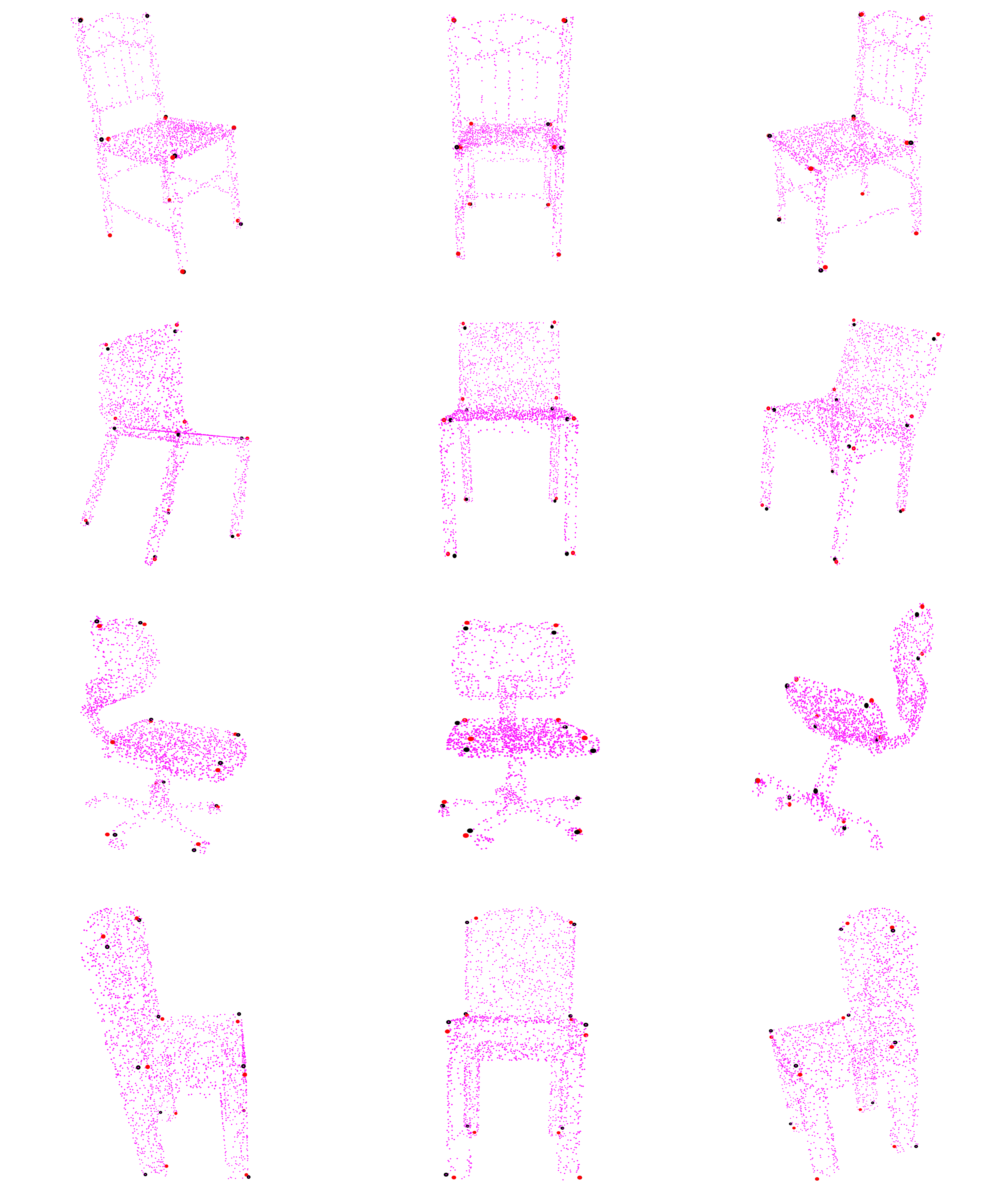}
  \vspace{0.01cm}
\caption{Visualization of predicted (red) and ground truth (black) keypoints. Each row represents a separate object from different viewpoints. Zoom in for details. }
\label{fig:keypoints}
\end{figure*}

\subsection{3D Keypoint Prediction}



We also consider the task of 3D keypoint prediction.
We use the training data provided by \cite{yi2017syncspeccnn} which includes 10 keypoints for each object in ShapeNet's chair category.  
We pre-train a model for rotation prediction and fine-tune it for keypoint regression. More specifically, we first train a rotation prediction model (on all categories) and then modify its final layers to regress $10$ keypoints for each input shape. 
The regression network is initialized with weights of the rotation prediction model and then fine-tuned with chamfer loss between predicted and ground truth keypoints. 
At inference time, we map each regressed point to its nearest neighbor point in the input point cloud. 

PCK\footnote{Percentage of Correct Correspondences} curves for our approach 
are shown in Figure~\ref{fig:pck}, and are compared with~\cite{yi2017syncspeccnn} and \cite{huang2013fine}. We observe that our method outperforms competing methods for all error thresholds. 
We also plot PCK curves for various sizes of the training set and see that our model achieves a good performance even when trained on a subset of training data. 
Note that compared with the shape classification task, keypoint prediction relies more on local and part-level information. 
Our results indicate that our self-supervised approach can transfer useful local and global information. 
We visualize the keypoints in Figure~\ref{fig:keypoints}. We observe that predicted and ground truth keypoints are close to each other, and in many cases they overlap. 







\section{Conclusion and Future Work}

We present a self-supervised approach for learning representations of 3D point clouds. A supervised signal for the pretext task of rotation prediction is created to learn representations that are useful for solving target tasks. We devise models for learning proxy and downstream tasks, and demonstrate results on shape classification and 3D keypoint prediction.   
The results indicate that our approach achieves superior  performance to competing methods. 
Each self-supervised learning method learns features tailored for its own proxy task. We demonstrate that features from different models can complement one another and outperform each individual model.  
Finally, we show that only a small subset of the training data suffices to achieve high accuracy on shape classification, eliminating the need for annotating massive 3D datasets. 

While our work focuses on point clouds, self-supervised learning can also be applied to other 3D representations such as meshes, voxel grids and implicit functions. Each representation has its own merits and disadvantages. Studying self-supervision for different representations is relatively under-explored and devising effective proxy tasks for various 3D representations is a promising venue for future work.  
We can also consider combining self-supervised techniques for different representations. 
Finally, evaluating our model on other downstream tasks is another direction for future research.     

{\small
\bibliographystyle{ieee}
\bibliography{egbib}
}

\end{document}